# Multimodal learning of melt pool dynamics in laser powder bed fusion

Satyajit Mojumder[1*], Pallock Halder[1], Tiana Tonge[1]

[1]School of Mechanical and Materials Engineering, Washington State University, Pullman, WA 99164, USA

**Abstract:**

While multiple sensors are used for real-time monitoring in additive manufacturing, not all provide practical or reliable process insights. For example, high-speed X-ray imaging offers valuable spatial information about subsurface melt pool behavior but is costly and impractical for most industrial settings. In contrast, absorptivity data from low-cost photodiodes correlate with melt pool dynamics but is often too noisy for accurate prediction when used alone. In this paper, we propose a multimodal data fusion approach for predicting melt pool dynamics by combining high-fidelity X-ray data with low-fidelity absorptivity data in the Laser Powder Bed Fusion (LPBF) process. Our multimodal learning framework integrates convolutional neural networks (CNNs) for spatial feature extraction from X-ray data with recurrent neural networks (RNNs) for temporal feature extraction from absorptivity signals, using an early fusion strategy. The model is further used as a transfer learning model to fine-tune the RNN model that can predict melt pool dynamics only with absorptivity, with greater accuracy compared to the multimodal model. Results show that training with both modalities significantly improves prediction accuracy compared to using either modality alone. Furthermore, once trained, the model can infer melt pool characteristics using only absorptivity data, eliminating the need for expensive X-ray imaging. This multimodal fusion approach enables cost-effective, real-time monitoring and has broad applicability in additive manufacturing.

**\*Corresponding author:** Satyajit Mojumder ( satyajit.mojumder@wsu.edu)

# 1. Introduction

Metal additive manufacturing (AM) process, particularly laser powder bed fusion (LPBF) has demonstrated tremendous potential in the production of highly complex and customized components for applications across aerospace, biomedical, and automotive industries [1,2]. LPBF operates by selectively melting fine layers of metal powder using a laser source, building up parts layer-by-layer[3–5]. The ability to produce intricate geometries and a reduction in material waste have positioned LPBF as a key technology in the shift toward digital manufacturing. However, despite its advantages, LPBF suffers from issues of process variability, sensitivity to small changes in process parameters, and lack of in-situ quality assurance [6–9]. These challenges can result in defects such as porosity, lack of fusion, crack formation or residual stress accumulation, which compromise part integrity. Therefore, achieving consistent part quality and reliability requires robust monitoring and control strategies during the build process [10–12]. To address these quality and reliability concerns, researchers and industry practitioners have implemented a wide range of sensing technologies to monitor LPBF processes in real time. Common modalities include high-speed optical imaging, infrared thermography, photodiode-based absorptivity sensing, X-ray imaging, and acoustic emission monitoring [13–16]. Each sensor type captures a unique facet of the physical process: for instance, infrared cameras record melt pool temperatures, while photodiodes track the dynamic absorptivity of the laser-material interaction[17–20]. More recently, high-speed X-ray imaging systems have provided detailed internal views of the melt pool and vapor cavity behavior, offering insight into subsurface phenomena [21,22]. Despite the richness of data collected, one of the key challenges lies in integrating this information coherently. The sheer volume, heterogeneity, lack of co-registered in-situ and ex-situ process dataset and differing spatial-temporal resolutions of sensor outputs make multimodal data fusion both necessary and complex [23,24]. If properly fused, these datasets can offer a more comprehensive view of the process than any single sensor alone, enabling more accurate detection of anomalies and prediction of melt pool characteristics [6,25].

Multimodal data fusion refers to the integration of information from different sensor modalities to improve understanding, decision-making, and system performance [26,27]. While the concept is well-established in fields such as robotics, surveillance, and remote sensing, its application in additive manufacturing is emerging and holds substantial potential [28–30]. In LPBF, individual sensor modalities are often limited by noise, occlusion, or incomplete representation of the process. For example, thermal cameras may not capture subsurface flaws, and acoustic sensors may suffer from signal interference. Combining their outputs can mitigate such weaknesses, enhance robustness, and reduce false-positive rates in defect detection [31,32]. Additionally, multimodal fusion enables the extraction of latent features that may only emerge when data from different sources are interpreted jointly. This is particularly useful in complex physical systems like LPBF, where thermal, optical, and mechanical events are intricately coupled. Consequently, fusion methods can provide richer process characterization and lay the groundwork for closed-loop control in AM systems.

Despite its promise, multimodal fusion in additive manufacturing remains a developing area. Most current research focuses on unimodal or, at best, bimodal sensing approaches, where data streams are often analyzed independently or combined heuristically [33]. These methods tend to lack the scalability and adaptability needed for real-world deployment. Furthermore, handcrafted feature engineering and domain-specific rules often limit generalizability across different machines, materials, or part geometries [30,34]. Recent progress in deep learning and sequence modeling has opened new avenues for data-driven multimodal fusion, offering the potential for automatic feature extraction, flexible architecture design, and improved inference performance [35–37]. However, the effective deployment of such models in AM settings still faces challenges, particularly in aligning heterogeneous data sources, managing missing modalities during inference, and achieving real-time performance.

This paper addresses these challenges by proposing a novel multimodal deep learning framework that fuses X-ray imaging data with absorptivity time-series data to predict melt pool dimensions during LPBF process for a spot laser. We leverage a Convolutional Neural Network (CNN) to extract spatial features from high-speed X-ray images and a Recurrent Neural Network (RNN) to model temporal patterns in absorptivity signals for individual modality. An early fusion of these modalities is performed during the training phase, enabling the model to learn a joint representation of the melt pool dynamics. Further, the multimodal model is used as a transfer learning model to fine tune the RNN model. At inference time, the transfer learning model operates solely on absorptivity data—an easily deployable and real-time sensor—while retaining the performance benefits of X-ray-informed training. This approach enables high-quality predictions in the absence of expensive or impractical sensor inputs during deployment. By designing the fusion architecture to decouple training and inference modalities, we offer a practical path toward robust, real-time melt pool monitoring in LPBF. Our experimental results demonstrate improved prediction accuracy over single-modality models and validate the feasibility of training with rich multimodal data while deploying in leaner, sensor-constrained environments.

## 2. Methodology
### 2.1 Data collection and preprocessing

The dataset used in this study was obtained from the NIST Public Data Repository, specifically from the 2022 Asynchronous AM-Bench Challenge [38]. The data was generated through a series of controlled experiments designed to investigate the relationship between energy absorption and melt pool geometry during laser-based additive manufacturing of Ti-6Al-4V alloy. Two sensing modalities were used: high-speed X-ray imaging and integrating sphere radiometry (ISR), allowing for simultaneous observation of melt pool dynamics and laser energy absorption. The details of the experiment are described in these [39,40].

The melt pool boundary was first identified through visual inspection, with visible boundaries in selected dataset images manually annotated using a purple line. From 148 spot laser images were annotated for melt pool features such as melt pool width and depth, and keyhole width and depth, following this [41]. Preprocessing involved loading grayscale images, normalizing pixel values to [0, 1], rotating them by 7° to align the material interface horizontally and the melt pool/keyhole vertically, and cropping to ensure geometric consistency. Similarly, the absorptivity data for the spot laser were collected and co-registered with the X-ray images. The absorptivity data was normalized using a MinMaxScaler for consistency. A sample melt pool image of the X-ray with key melt pool features is presented in Figure 1.

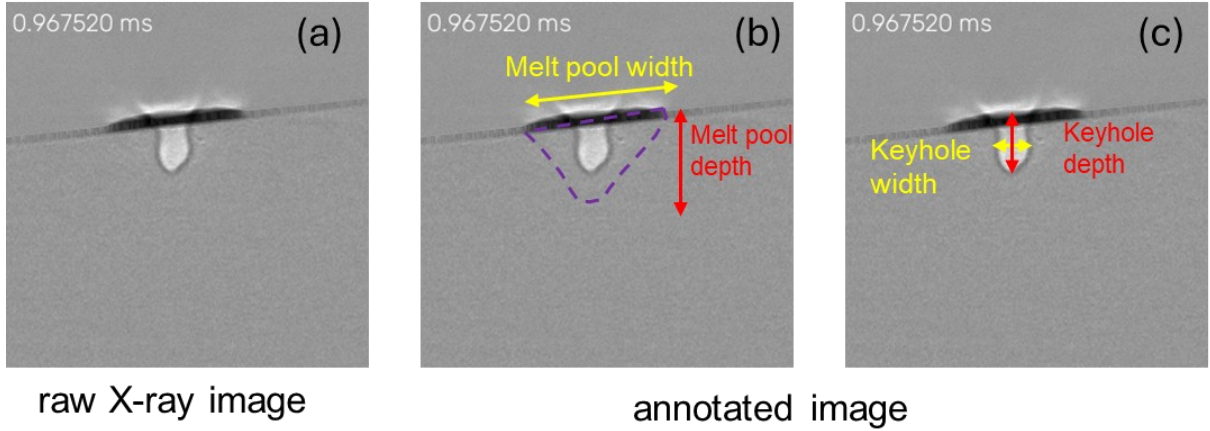

Figure 1: (a) Raw X-ray image of melt pool formation with a spot laser. The X-ray image is processed to identify the melt pool (b) and the keyhole contours (c) to determine the key melt pool features, such as melt pool and keyhole width and depth.

## 2.2 Model architecture

The primary objective of this work is to train a deep learning model to predict key melt pool characteristics—specifically, the width and depth of both the melt pool and the keyhole—using high-fidelity X-ray images and low-fidelity absorptivity data. To first understand the contribution of each data source, we begin with single-modality learning, training separate models on each dataset. The X-ray images are processed using a Convolutional Neural Network (CNN) architecture, while the time-dependent absorptivity data is modeled using a Long Short-Term Memory (LSTM) Recurrent Neural Network (RNN) architecture. These single-modality models establish a baseline for subsequent multimodal learning.

In the multimodal approach, the CNN and RNN architectures are integrated to jointly learn melt pool features from both data sources. An early-fusion strategy is employed by implementing a combined CNN-RNN architecture that combines the processed X-ray images with the corresponding absorptivity data. Recognizing the high cost and limited availability of X-ray imaging, we further leverage the trained multimodal model in a transfer learning framework, fine-tuning the RNN to make predictions using only absorptivity data. This enables accurate inference without requiring X-ray inputs during deployment.

### 2.2.1 CNN for melt pool prediction from X-ray data

This is a single modality model and only uses the X-ray images as input and the melt pool features as target. A CNN architecture was implemented to capture the complex features of the melt pool and keyhole geometry from X-ray images. The input images were gray-scaled, resized to 128 × 128 pixels and normalized to [0,1], paired with their corresponding width-to-depth ratio labels. The architecture comprised four convolutional blocks with progressively increasing filters (32, 64, 128, 256) using 3×3 kernels and Rectified Linear Unit (ReLU) activations, each followed by batch normalization and 2×2 max pooling to stabilize learning and reduce dimensionality. A global average pooling layer was used to aggregate spatial features, which were then passed through a fully connected dense layer with 512 neurons and ReLU activation, alongside a dropout (0.2) for regularization. The final output layer was a single linear neuron predicting the melt pool or keyhole width-to-depth ratio. The network was trained with the Adam optimizer and mean squared error (MSE) loss over 10,000 epochs with a batch size of 8 and evaluated on an 80:20 train-test split.

### 2.2.2 RNN for melt pool prediction from absorptivity data

This single-modality model uses absorptivity data as input and melt pool features as the prediction target. To capture the temporal dependencies between laser absorptivity and the evolution of the melt pool or keyhole, an RNN framework was implemented using a bidirectional long short-term memory (Bi-LSTM) architecture. The input consisted of absorptivity measurements, reshaped into a three-dimensional tensor of the form [N, T, F], where N is the number of samples, T=1 denotes the sequence length (single time step per input), and F=1 represents the single absorptivity feature. Although a unit-length sequence was employed in this study, the model design is readily extensible to longer temporal windows. The recurrent backbone comprised two stacked Bi-LSTM layers with 128 hidden units in each direction, with ReLU activations and He-normal weight initialization. Each Bi-LSTM layer was followed by batch normalization and a dropout layer (rate = 0.05) to mitigate overfitting and stabilize convergence. To enhance feature representation, a multi-head self-attention module (4 heads, key dimension = 16) was applied to the Bi-LSTM outputs, enabling the network to model long-range dependencies within the temporal domain. The attention-enhanced features were subsequently projected through a feed-forward network consisting of four fully connected layers with 128 neurons each, ReLU activations, and dropout regularization (rate = 0.05). The final regression layer was a single linear neuron, outputting the predicted melt pool width-to-depth ratio.

The network was trained using the Adam optimizer with a fixed learning rate of $1\times10^{-4}$, optimizing the mean squared error (MSE) loss function. Training was conducted for a maximum of 10,000 epochs with a batch size of 32, reserving 20% of the dataset for validation. To ensure robust convergence, early stopping (patience = 80, with best-weight restoration) and adaptive learning rate scheduling by ReduceLROnPlateau (factor = 0.5, patience = 30, minimum learning rate = $1\times10^{-6}$) were applied. This architecture, by integrating Bi-LSTM units with an attention mechanism, provides a scalable and generalizable framework for capturing both short and long-range correlations between absorptivity signals and melt pool morphology.

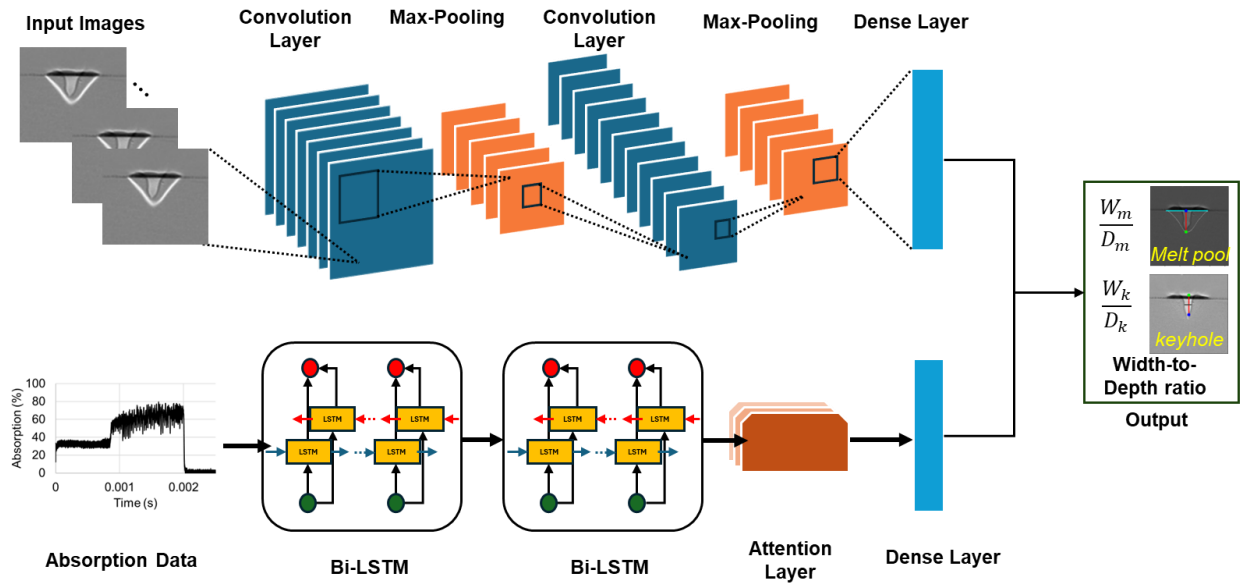

Figure 2: Model architecture for the melt pool features prediction from multimodal X-ray images and absorptivity data

### 2.3.3 Multimodal CNN-RNN fusion for melt pool prediction

To jointly use spatial and temporal information, a multimodal model comprising CNN and RNN networks, was developed by integrating CNN features extracted from X-ray image data with recurrent features derived from absorptivity measurements (Fig. 2). The CNN branch processed grayscale images of size 128 × 128 pixels through two convolutional blocks (Conv2D with 32 filters, 3×3 kernels, ReLU activations followed by 2×2 max pooling) and a flattening layer, concluding with a fully connected layer of 64 neurons (ReLU). In parallel, the absorptivity signal, reshaped as a single-feature input vector, was processed through a dense layer of 32 neurons (ReLU) to provide a representation of temporal absorption dynamics. The outputs of the CNN and RNN branches were then fused by concatenation, enabling the model to learn melt pool features from image and absorptivity data. This merged representation was passed through an additional fully connected layer of 64 neurons (ReLU) before being mapped to the final linear regression head, which predicted the melt pool or keyhole width-to-depth ratio. The combined network was trained using the Adam optimizer with mean squared error (MSE) loss over 5,000 epochs with a batch size of 32. Model performance was evaluated using mean absolute error (MAE) and coefficient of determination ($R^2$), and both true and predicted width-to-depth ratios were compared across the temporal axis.

### 2.2.4 Transfer learning model

To reduce model complexity while retaining predictive accuracy, a knowledge distillation framework was used in which a multimodal CNN–RNN fusion network served as the teacher model, and a lightweight RNN architecture acted as the student model [42,43]. The teacher model (Section 2.3.3) combined convolutional features from X-ray images with process features from absorptivity signals through concatenation, producing high-fidelity predictions of the melt pool width-to-depth ratio. These predictions were then leveraged as supervisory signals to guide the training of the student model, which was restricted to absorptivity data alone.

The motivation for this framework comes from the practical limitations of in-situ X-ray imaging. While X-ray diagnostics provide highly detailed information about melt pool dynamics, they are extremely costly and are not scalable to large-scale or industrial additive manufacturing processes. In contrast, absorptivity data can be obtained using relatively inexpensive and non-intrusive sensors. Therefore, if a lightweight RNN trained on absorptivity alone can reproduce the predictions of a fused CNN–RNN model that relies on expensive X-ray data, the approach offers a more cost-effective and scalable pathway for monitoring melt pool behavior.

For transfer learning, the absorptivity sequences were constructed using a fixed sequence length of 5, reshaping the input into [N, T, F] with T=5 (time steps) and F=1 (absorption). The student architecture consisted of a single LSTM layer with 32 hidden units, followed by a fully connected layer of 64 neurons with ReLU activation, and a final linear regression output layer to predict the width-to-depth ratio. Unlike the teacher network, which integrated both spatial and temporal modalities, the student model relied solely on the temporal absorptivity sequences but learned to approximate the teacher's outputs through distillation. Training was performed with the Adam optimizer and mean squared error (MSE) loss, using mini-batches of size 8 and up to 10,000 epochs. To ensure alignment, teacher predictions were first generated for the full dataset and stored, after which the RNN student was trained to minimize the discrepancy between its predictions and the teacher's outputs, while also being evaluated against the true experimental labels. Model performance was quantified using MAE and $R^2$ score.

## 3. Results

This section presents the predictive capabilities and comparative performance of the individual models. The analysis begins with single-modality results for each architecture, followed by multimodal performance, highlighting the gains achieved through data fusion. Quantitative metrics such as MAE and $R^2$ are reported alongside qualitative comparisons, including visual overlays of predicted versus actual melt pool and keyhole dimensions. These results provide a clear basis for evaluating model accuracy, generalization, and the effectiveness of the proposed learning strategies.

### 3.1 X-ray-only modality with CNN

This single-modality model uses X-ray images as input and melt pool features as the prediction target. Figure 3 presents a comparison between the CNN model predictions and the corresponding X-ray measurements ground truth for both melt-pool and keyhole width-to-depth ratios. The predicted trends closely follow the experimental ground truth across the entire temporal range of the spot laser dynamics. For the melt-pool case (Figure 3a), the CNN captures the sharp fluctuations in the ratio during the initial transient stage. Similarly, for the keyhole case (Figure 3b), the network successfully reconstructs both the rapid drop immediately after laser initiation and the subsequent low-ratio steady state.

The evaluation metrics further confirm the high predictive accuracy of the model. For the melt-pool width-to-depth ratio, the CNN achieves a mean absolute error (MAE) of 0.0477 and an $R^2$ score of 0.9633, indicating excellent agreement with experimental data. For the keyhole width-to-depth ratio, the model performs comparably, with an MAE of 0.0401 and an $R^2$ score of 0.9698. A high accuracy in $R^2$ score demonstrates the high-fidelity nature of the X-ray images to preserve the melt pool features. The CNN model effectively captures the melt pool features by learning these features of melt pool and keyhole geometries.

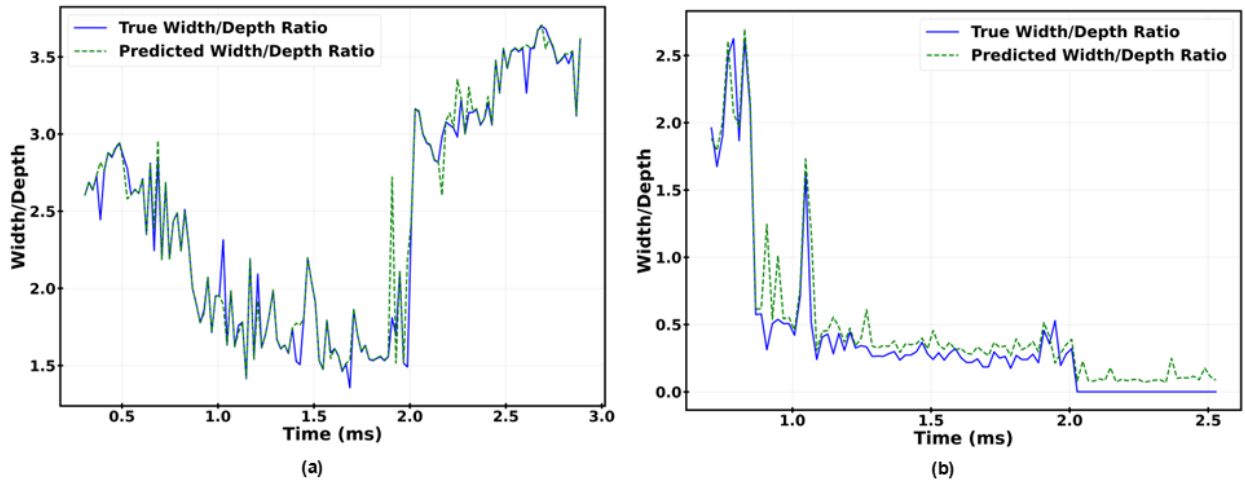

Figure 3: Comparison of CNN model predictions with X-ray measurements for: (a) melt-pool width-to-depth ratio and (b) keyhole width-to-depth ratio

## 3.2. Absorptivity modality with RNN

The single-modality RNN model uses absorptivity data as input and melt pool features as the prediction target. Figure 4 shows the performance of the RNN model trained exclusively on absorptivity data for predicting the melt-pool and keyhole width-to-depth ratios. Compared to the CNN model (section 3.1) with X-ray inputs, the RNN demonstrates lower fidelity in capturing the fine temporal variations of the ratios. For the melt-pool case (Figure 4a), the model can approximate the overall trend but fails to fully reproduce the sharp oscillations and rapid local fluctuations observed in the ground truth. Similarly, for the keyhole case (Figure 4b), the RNN predictions capture the general decreasing trend but cannot estimate the steep initial transitions. The quantitative results also highlight this performance gap. For the melt-pool width-to-depth ratio, the RNN achieves a mean absolute error (MAE) of 0.2063 and an $R^2$ score of 0.8695. For the keyhole width-to-depth ratio, the error further increases (MAE = 0.2341) while the $R^2$ score drops to 0.8078, indicating limited ability to replicate the true dynamics. The performance of the RNN model reflects the low-fidelity and noisy nature of the absorptivity data. Despite this, the model successfully captures the key melt pool dynamics during the steady-state region, while its predictive accuracy decreases during the initial melt pool formation and after the laser is turned off.

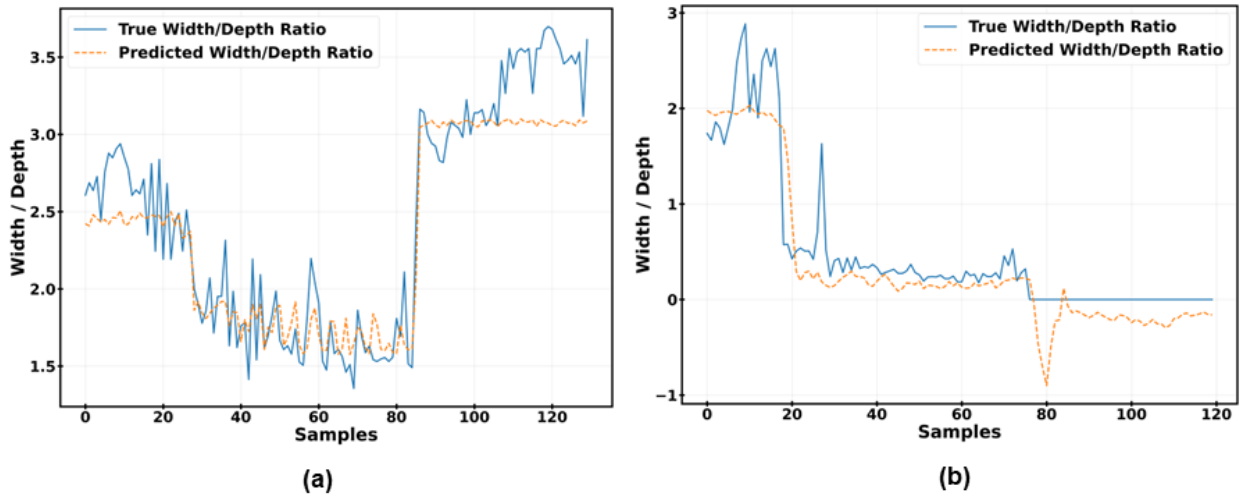

Figure 4: Comparison of RNN model predictions with absorption data for: (a) melt-pool width-to-depth ratio and (b) keyhole width-to-depth ratio

## 3.3 Multimodality with combined CNN-RNN

The combined CNN–RNN framework integrates both X-ray image features (CNN) and absorptivity signals (RNN), aiming to utilize the complementary strengths of each modality. Figure 5 compares the model predictions against X-ray ground truth for both melt-pool (Figure 5a) and keyhole width-to-depth ratios (Figure 5b).

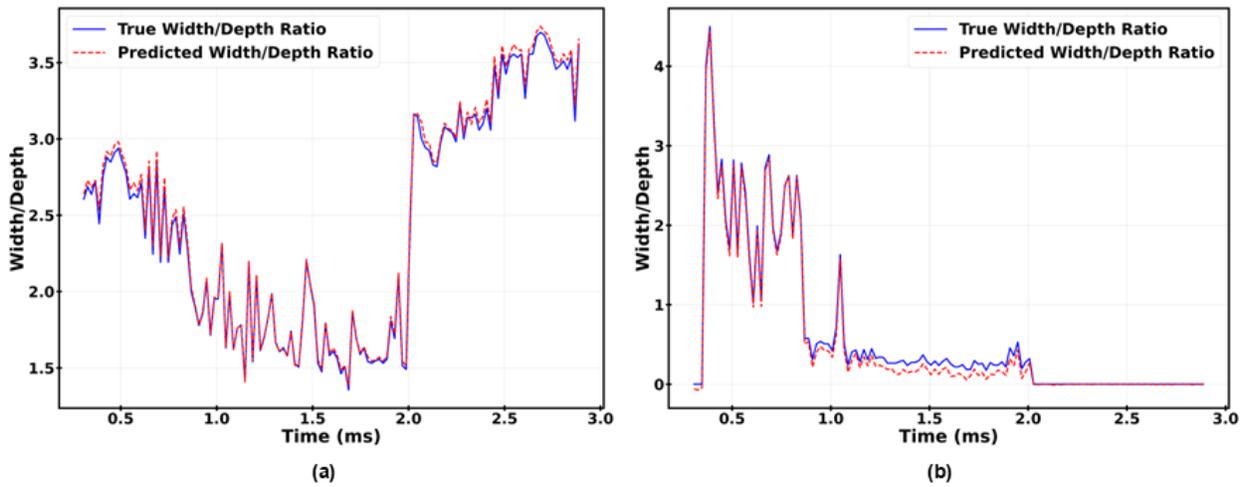

Figure 5: Comparison of Combined CNN-RNN model predictions with X-ray measurements for: (a) melt-pool width-to-depth ratio and (b) keyhole width-to-depth ratio

The quantitative results confirm the accuracy of the combined approach. For the melt-pool width-to-depth ratio, the model achieves an MAE of 0.1744 with an $R^2$ score of 0.9241. For the keyhole case, the predictive accuracy is even stronger, with an MAE of 0.0886 and an exceptionally high $R^2$ score of 0.9842, demonstrating near-perfect alignment with experimental measurements. The combined CNN–RNN model yields a marginally lower $R^2$ score for melt pool width-to-depth ratio prediction compared to the CNN-only model yet achieves a substantially higher score than the RNN-only counterpart. These findings indicate that the multimodal architecture not only enhances predictive accuracy relative to the absorptivity-driven

RNN but also surpasses the X-ray-based CNN in modeling transient keyhole dynamics. Importantly, this improvement means that accurate predictions can be achieved even when relying solely on low-fidelity absorptivity data, reducing dependence on costly high-fidelity X-ray imaging and enabling more practical, scalable deployment in real manufacturing environments.

**3.4 Transfer learning model**

Multimodal learning offers superior predictive performance compared to single-modality models; however, the acquisition and processing of multimodal datasets can be resource-intensive and time-prohibitive. To address this, we employ a transfer learning approach that leverages knowledge from multimodal training for single-modality prediction. Specifically, the CNN–RNN model was trained with both X-ray and absorptivity inputs, then tested using only the absorptivity (RNN) channel to evaluate the effectiveness of the transferred representations. Figure 6 shows the comparison between predicted and ground-truth width-to-depth ratios for both melt-pool and keyhole cases. The predicted curves follow the experimental trends more closely than the absorptivity-only RNN model, with improved reproduction of transient fluctuations and better alignment during both steady and unstable regimes.

The performance metrics for the melt-pool width-to-depth ratio, the transfer model achieves an MAE of 0.1658 and an $R^2$ score of 0.9336, which represents a substantial improvement over the absorptivity-only RNN model. Similarly, for the keyhole width-to-depth ratio, the transfer model reaches an MAE of 0.1588 and an $R^2$ score of 0.9360, markedly higher than the RNN-only performance. These results demonstrate that the teacher–student transfer strategy successfully distills knowledge from multimodal training into the unimodal student model. Importantly, this enables accurate predictions using only absorptivity data—bypassing the need for costly X-ray imaging—while retaining much of the predictive fidelity of the full multimodal CNN–RNN.

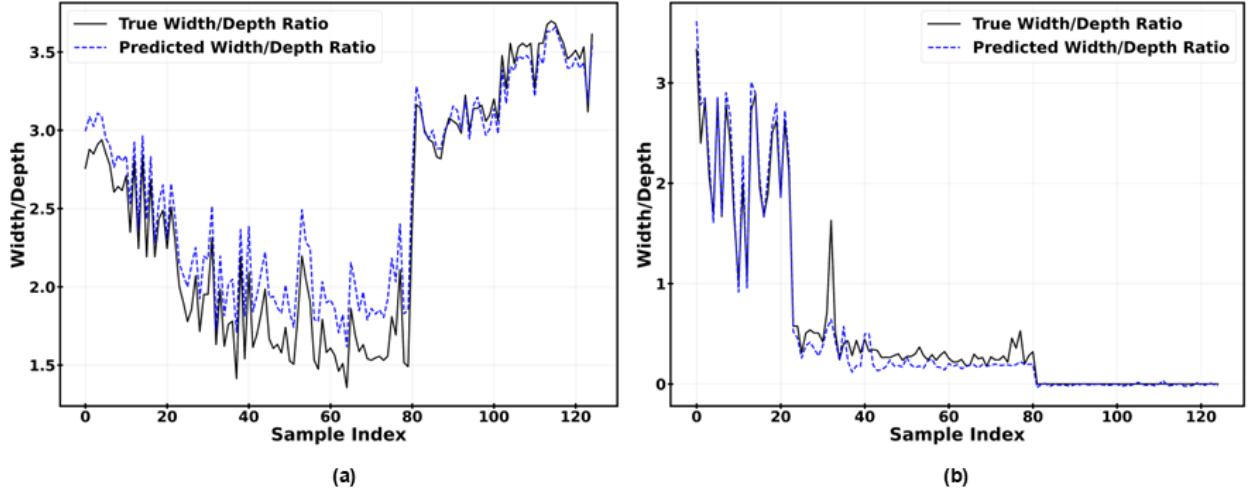

Figure 6: Comparison of Transfer CNN-RNN model predictions with absorption data only for: (a) melt-pool width-to-depth ratio and (b) keyhole width-to-depth ratio

### 3.5. Model performance evaluation

The comparative evaluation of the four models—X-ray-only CNN, absorptivity-only RNN, combined CNN–RNN, and transfer CNN–RNN—highlights the trade-off between predictive accuracy and data acquisition cost. Table 1 (summary table of metrics across models) consolidates the results for both melt-pool and keyhole width-to-depth ratios. The X-ray-only CNN (Section 3.1) achieved the highest accuracy when relying solely on image-based features, with $R^2$ scores above 0.96 for both melt-pool and keyhole ratios. This demonstrates the strong capability of CNNs to capture fine spatial features from high-resolution imaging. However, its dependency on costly X-ray imaging limits scalability for in-situ process monitoring.

The absorptivity-only RNN (Section 3.2) provided a low-cost alternative as it relies only on process absorptivity signals. While still capable of approximating overall trends ($R^2 = 0.87$ for melt-pool and 0.81 for keyhole), it showed difficulty in capturing sharp transients and fine-scale variations, leading to higher MAE compared to the CNN. The combined CNN–RNN fusion model (Section 3.3) outperformed both unimodal approaches, achieving an $R^2$ of 0.9241 for melt-pool and an exceptionally high 0.9842 for keyhole predictions. By integrating image-based spatial features with temporal absorption signals, this model successfully leverages the complementary information, resulting in robust and accurate predictions across both regimes. Finally, the transfer CNN–RNN (Section 3.4) demonstrated the effectiveness of knowledge distillation. Trained with multimodal inputs but tested on absorptivity alone, the model achieved R² scores of 0.9336 (melt-pool) and 0.9360 (keyhole)—substantially better than the absorptivity-only RNN and approaching the accuracy of the multimodal fusion model. This result underscores the practical value of transfer learning in enabling low-cost deployment while preserving high predictive fidelity.

Table 1: Summary of evaluation metrics across all models

| Model | Melt-pool MAE | Melt-pool $R^2$ | Keyhole MAE | Keyhole $R^2$ |
|---|---|---|---|---|
| CNN (X-ray only) | 0.0477 | 0.9633 | 0.0401 | 0.9698 |
| RNN (Absorptivity only) | 0.2063 | 0.8695 | 0.2341 | 0.8078 |
| Combined CNN-RNN | 0.1744 | 0.9241 | 0.0886 | 0.9842 |
| Transfer CNN-RNN | 0.1658 | 0.9336 | 0.1588 | 0.9360 |

## 4. Discussion

The results of this study demonstrate the value of multimodal data fusion for improving melt pool prediction in the LPBF process. By combining X-ray and absorptivity data during training, our model leverages complementary sources of information that capture both spatial and temporal characteristics of the melt pool. The X-ray images provide detailed subsurface views of melt pool morphology and keyhole dynamics, while the absorptivity signals capture real-time fluctuations in the laser-material interaction. This fusion acts as a feature augmentation mechanism, enabling the model to learn a richer and more discriminative representation of the process state than either modality can provide alone. As a result, the fused model exhibits improved prediction accuracy and robustness compared to unimodal baselines.

A particularly impactful aspect of this approach is the decoupling of training and inference modalities. While high-speed X-ray imaging offers critical insights into melt pool behavior, it is costly, difficult to scale, and impractical for deployment in most industrial LPBF environments. The proposed architecture mitigates this limitation by using X-ray data during training to guide feature learning, then relying solely on absorptivity data—an inexpensive and easy-to-integrate sensor—for inference. This significantly lowers the barrier for real-time implementation and makes the model viable in practical manufacturing settings where only low-cost or non-intrusive sensors are available.

Beyond absorptivity, the multimodal fusion framework developed here is extensible to other sensor combinations. Acoustic emission data, ultrasonic monitoring, high-speed optical cameras, and melt pool visible-light imaging can be integrated into similar architectures. These sensors, like X-rays and photodiodes, capture different aspects of the melt pool state and defect formation, and their integration through deep learning offers a path toward comprehensive in-situ process monitoring. The model-agnostic nature of convolutional and recurrent layers means that with appropriate preprocessing and temporal alignment, these modalities can be added, replaced, or removed based on availability or cost constraints.

Despite these benefits, one limitation of the current work is the reliance on a controlled and relatively small dataset. The X-ray and absorptivity data were collected under a limited set of process parameters and machine conditions, which may restrict the generalizability of the trained model. In real-world deployments, process variability across machines, materials, and environmental conditions introduces domain shifts that the current model may not fully handle. Future work should explore domain adaptation techniques and more diverse datasets to improve model robustness.

From an engineering standpoint, the approach balances prediction accuracy with deploy ability. Training with expensive, high-fidelity sensors and inferring with low-cost alternatives allows manufacturers to improve quality assurance without significantly increasing monitoring costs. This paradigm is particularly important for scaling additive manufacturing to high-volume production environments, where real-time monitoring must be efficient, non-intrusive, and cost-effective. However, challenges remain, particularly in scaling the model to new materials, geometries, or environmental conditions. Adaptive learning strategies, sensor redundancy, and uncertainty quantification may help address these challenges and make multimodal fusion a core part of next generation AM quality control systems.

## 5. Conclusion

In this study, we developed and evaluated deep learning frameworks for predicting melt pool and keyhole geometries using both high-fidelity X-ray images and low-fidelity absorptivity data. Single-modality models established the baseline capabilities of each data source, while the multimodal CNN–RNN fusion approach demonstrated clear performance gains by leveraging complementary spatial and temporal features. Notably, the multimodal model not only outperformed the absorptivity-only RNN but also surpassed the X-ray-only CNN in capturing transient keyhole dynamics. Through transfer learning, we further showed that knowledge gained from multimodal training can be effectively applied to single-modality absorptivity data, preserving much of the predictive accuracy while eliminating the need for the costly X-ray imaging requirement of the multimodal model. This capability significantly enhances the deployability of the approach for real-time, in-situ monitoring in manufacturing environments. The flexibility of the fusion framework makes it adaptable to other sensor types, such as acoustic or ultrasonic monitoring, broadening its applicability to a wide range of additive manufacturing processes.

**Acknowledgements:**

The authors gratefully acknowledge the WSU start-up grant for supporting this research. They also thank Mr. Apurba Sarker and Mr. M. Abrar Muhit for their valuable discussions through the WSU SUPREME program for this work.

**Data Availability:**

Data and the codes will be available upon reasonable request to the corresponding author.

**Declaration of interests:**

The authors declare that they have no known competing financial interests or personal relationships that could have appeared to influence the work reported in this paper.